\documentclass{article}

\usepackage[final]{neurips_2019}

\usepackage{xcolor}
\usepackage[utf8]{inputenc} 
\usepackage[T1]{fontenc}    
\usepackage{hyperref}       
\usepackage{url}            
\usepackage{booktabs}       
\usepackage{amsfonts}       
\usepackage{amsmath}       
\usepackage{nicefrac}       
\usepackage{microtype}      
\usepackage{graphicx}

\title{Phenotyping of Clinical Notes with Improved Document Classification Models Using Contextualized Neural Language Models}
\author{
    Andriy Mulyar$^1$\thanks{Work performed as a visiting student at Johns Hopkins University.} \quad Elliot Schumacher$^2$ \quad Masoud Rouhizadeh$^3$ \quad Mark Dredze$^2$\\ 
    $^1$Department of Computer Science, Virginia Commonwealth University \\
    $^2$Department of Computer Science, Johns Hopkins University\\
    $^3$Institute for Clinical and Translational Research , Johns Hopkins University
}

\begin{document}

\maketitle

\begin{abstract}
Clinical notes contain an extensive record of a patient's health status, such as smoking status or the presence of heart conditions. However, this detail is not replicated within the structured data of electronic health systems.  Phenotyping, the extraction of patient conditions from free clinical text, is a critical task which supports a variety of downstream applications such as decision support and secondary use of medical records.  Previous work has resulted in systems which are high performing but require hand engineering, often of rules \cite{uzuner2008identifying, uzuner2009recognizing}.  Recent work in pretrained contextualized language models \cite{DBLP:conf/naacl/DevlinCLT19} have enabled advances in representing text for a variety of tasks.  We therefore explore several architectures for modeling phenotyping that rely solely on BERT representations of the clinical note, removing the need for manual engineering. We find these architectures are competitive with or outperform existing state of the art methods on two phenotyping tasks.

\end{abstract}







\section{CORRECTION}
{\em Added September 13, 2020}

{\color{red}
The original paper was published in December 2019. After publication, we identified a bug in our code that resulted in an error in our reported results. This version of the paper corrects that error and clarifies some of our descriptions of the experiments.

Specifically:
\begin{itemize}
    \item We have updated the results for each architecture. The results are now lower. They  still beat the previous shared task results on smoking but not obesity.
    \item We removed the cross-validation experiments and focused on the held out evaluation.
    \item Updated the description in the methods section.
\end{itemize}

The original version is available on Arxiv for comparison.

}

\section{Introduction}
Electronic Health Record (EHR) systems contain a wealth of health information about patients, including structured data (e.g. demographics, medical codes, lab results) and unstructured text in the form of clinical notes. While structured data includes information that characterizes medical conditions of the patient, it often does not include numerous characteristics of interest that are typically contained in the clinical notes. For example, while a clinician may not code the smoking status of a patient, it can appear in the notes if it was discussed during the visit. Since the mention of smoking status appears in unstructured text, the phrasing can vary: ``The patient smokes regularly'' or ``He reports a history of smoking.''
Including these characteristics as structured data can support both improved
patient care and secondary use of medical records.

The task of automatic identification of specific phenomic traits within patients is called {\it phenotyping}. Phenotyping is critical to cohort selection, in which a study selects a population from an EHR system for further study, e.g. men over 50 who smoke. There have been multiple shared tasks, including identifying smoking status \cite{uzuner2008identifying} and obesity related co-morbidities \cite{uzuner2009recognizing}, which have produced high-performing systems for obesity \cite{ware2009natural, solt2009semantic} and smoking status \cite{clark2008identifying,Yao2018ClinicalTC}.
While these successes are promising, the uses of phenotyping are vast, for which an almost unlimited number of phenomic traits can be useful for patient care or cohort selection. Therefore, systems that require extensive preprocessing (e.g abbreviation and negation detection) or feature engineering (e.g. detection of temporal phrases) targeted at specific tasks may have limited utility in supporting a diverse range of phenomic traits. 

Recent work on contextualized neural language models \cite{DBLP:conf/naacl/DevlinCLT19,peters2018deep,dai2019transformer} has led to systems which construct representations of text data that can support many tasks, with task specific training data only needed to train a final prediction layer. These models have been applied in the clinical space \cite{DBLP:journals/corr/clinicalBERT}.
However, these models -- which require building representations across the entire input -- have only been used for sentences or short segments of text. Little work has applied these models to entire documents or entire clinical notes.

We develop a phenotyping system based on neural contextualized representations of language. We utilize a clinically fine-tuned \cite{DBLP:journals/corr/clinicalBERT} version of BERT \cite{DBLP:conf/naacl/DevlinCLT19}. BERT can only be applied directly to relatively short spans of text. Since the phenomic trait can be contained anywhere in the clinical note, we explore ways of combining BERT representations from multiple segments into a single document-level representation. Recent BERT based document classification architectures \cite{DBLP:journals/corr/abs-1904-08398} are not suited for clinical notes as they consider only the first few sentences of the text thus cannot capture information relayed further into the document. We evaluate our approach on two domains of phenomic traits (obesity co-morbidities and smoking) and find that our best approach to document level modeling outperforms previous state of the art systems on smoking.



\vspace{-.3cm}
\section{Phenotyping of Clinical Notes with BERT}
\begin{figure}
  \centering
  \includegraphics[width=\textwidth]{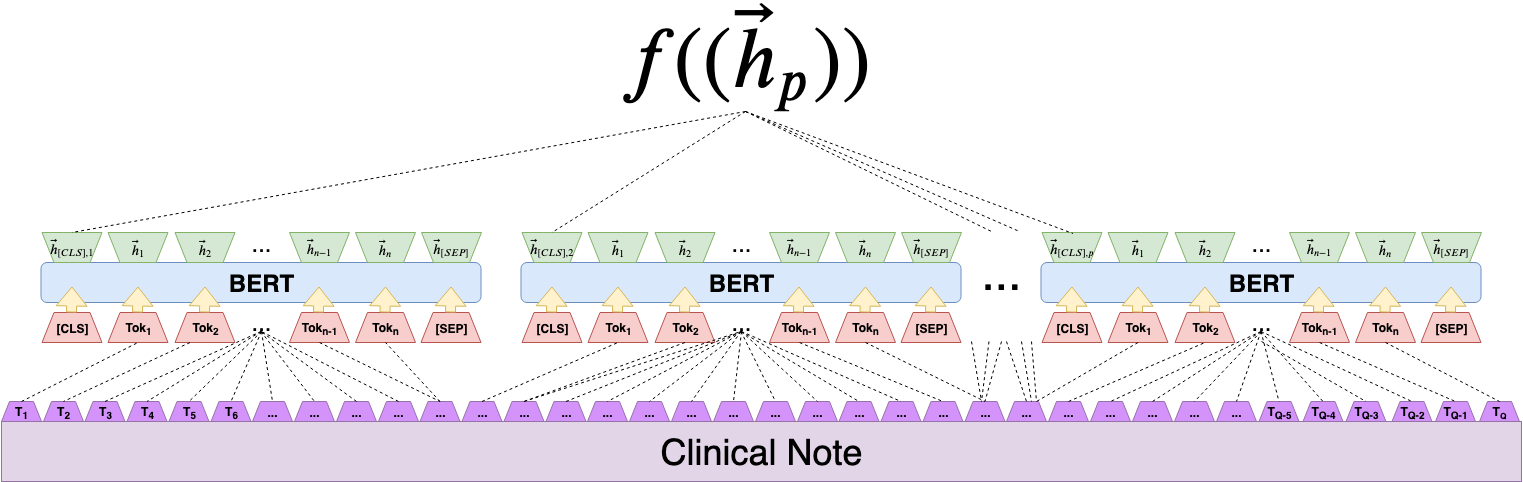}
  \caption{Our model architecture for a document classifier for phenomic traits based on repeated application of BERT to spans in a clinical note. Working bottom-up, a document is tokenized and chunked into spans according to the maximum BERT sequence length. An encoding function $f((\vec{h}_p))$ condenses the unrolled language model hidden state sequence  $(\vec{h}_p)$. Not shown, this encoded representation is then fed into a layer of perceptrons with sigmoid activation for classification.}
  \label{fig:architecture}
\end{figure}

We frame phenotyping as a classification problem, where for each phenomic trait for a clinical note our system produces a label, either binary or multi-class. Our classifier uses BERT to generate a sequence of representations of the text document. This representation sequence is condensed into a single document representation and then fed to our classifier. We first describe how BERT is applied to the document, and then describe several methods to construct a single document representation from the BERT output.


BERT considers a fixed length of text, e.g. $\ell = 512$ WordPieces/subwords, where tokens are subdivided using the WordPiece algorithm \cite{wu2016googles}.
BERT passes these subwords through its multiple transformer layers and produces both a per subword representation, as well as a representation meant to summarize the entire input: {\tt CLS}.
We divide the clinical document into chunks, where each chunk is of length $\ell$ (with the exception of possibly the last) and is passed into BERT. The result of this process is that we have a representation for each subword of input, as well as a {\tt CLS} representation for each text chunk. This process is shown in Figure \ref{fig:architecture}.



The next step requires the combination of the output from each BERT segment into a single document embedding. Previous approaches consider only the first two sentences as the basis for the representation \cite{Wu:2019rw}. However, since the phenomic trait can be contained anywhere in the document, we need to combine all {\tt CLS} embeddings into a single representation. Additionally, since clinical notes can be of arbitrary length, we need a general method to collapse this sequence of segment representations into a single representation. This combination function is represented in Figure \ref{fig:architecture} as $f((\vec{h}_p))$.


We consider four options for $f((\vec{h}_p))$. The input to each function is a sequence of the {\tt CLS} embeddings. Each element of the sequence itself summarizes the document segment that element encodes.
\begin{itemize}
    \item[] $f_{\text{mean}}$ :  A dimension-wise mean over all {\tt CLS} embeddings.
    \item[] $f_{\text{I}}$ : The identity function (a concatenation of all {\tt CLS} embedding).
    \item[] $f_{\text{Transformer}}$ : A dimension-wise max of the output sequence in the encoder layer of a Transformer \cite{vaswani2017attention}. 
    \item[] $f_{\text{LSTM}}$: The last hidden state of an LSTM run left-to-right over the {\tt CLS} embeddings. 

\end{itemize}



Each of these four architectures are identical up to the choice of encoding function $f$. In our proposed architectures (Figure \ref{fig:architecture}), we use an instance of ClinicalBERT \cite{DBLP:journals/corr/clinicalBERT} - a BERT language model fine-tuned over biomedical and clinical domain text. 
We use BERT's maximum word piece input length of $512$, using $510$ tokens
with a padding on each side.
The output of each {\tt CLS} combination function $f$ is linearly projected (via a layer of perceptrons) into the label space followed by a sigmoid activation. This entire architecture is trained on the available training set, which updates the parameters of the linear projection and if applicable $f$, e.g. the encoder layer of the transformer or the LSTM model parameters. 
During training, we randomly dropout \cite{srivastava2014dropout} weights with probability $.1$ in the projection, use binary cross entropy loss to estimate the target label distribution and only backpropagate weight updates in the last Transformer layer of BERT.

\vspace{-.3cm}
\subsection{Implementation details}

During training we utilize the BERT-base hidden size ($768$ dimensions) throughout all relevant internal hidden states in the $f_\text{LSTM}$ and $f_{\text{Transformer}}$ architectures and a dropout probability of $.1$ in the projection layer. During prediction we apply a given label if the sigmoid of the projections component corresponding to the label exceeds a threshold of $.5$; otherwise, the document does not receive the corresponding label. We trained all architectures on an NVIDIA Tesla M40 GPU. All architectures took approximately 24 hours of wall time (with negligible CPU computation) to converge to the performance reported in Table \ref{tab:results} on the N2C2 2006 training set and approximately 36 hours of wall time to converge on the N2C2 2008 training set.




\vspace{-.3cm}
\section{Evaluation}


We consider two clinical note phenotyping datasets released as part of shared tasks by N2C2 (formally named I2B2): 2006 Smoker Identification \cite{uzuner2008identifying} and 2008 Obesity Risk Factors \cite{uzuner2009recognizing}. The datasets are publicly available.\footnote{\url{https://portal.dbmi.hms.harvard.edu/data-sets/}}
Smoking consists of a single prediction task: select 
smoking status from four fine-grained options: past smoker, current smoker, non-smoker, and status unclear.
Obesity contains a label for obesity and 14 co-occurring morbidities (e.g. congestive heart failure), so a clinical note can have 0 or more applicable labels. For the obesity dataset we train and evaluate only using the ``intuitive judgments'' labels as these provide the largest annotation coverage over the data.


Each dataset contains a pre-defined training set identical to the data available to participants during the shared task competition period. We train each architecture over each respective training set, and report micro-averaged $F_1$ for each architecture on the evaluation set after $1000$ training epochs (a suitable number selected during development). We do not perform any additional hyper-parameter optimization since there is no development set. 
We include for comparison both the best system at the shared tasks and subsequent work with CNNs for these tasks.



\section{Results and Discussion}
\label{tab:results}
\begin{table}[h!]
  \caption{Phenotyping results (micro-averaged $F_1$) of our architectures trained on the respective shared task training sets and evaluated on the evaluation sets.}
  \centering
  \begin{tabular}{lcc}
    \toprule
     & \multicolumn{1}{c}{I2B2 2006: Smoking} & \multicolumn{1}{c}{I2B2 2008: Obesity}                   \\
    \cmidrule(r){2-2}
    \cmidrule(r){3-3}

    $f_{\text{mean}}$ & $\textbf{92.8}$  & $86.5$\\
    $f_{I}$     & $91.1$  & $82.9$ \\
    $f_{\text{Transformer}}$   & $58.4$ & $70.4$ \\
    $f_{\text{LSTM}}$     & $92.3$  & $83.1$ \\ \cmidrule(r){1-3}
    Shared Task $1^{\text{st}}$ Place     & $90.0$  & $95.0$\\
    Majority Label Baseline & $81.0$ & $74.4$ \\
    DocBert \cite{DBLP:journals/corr/abs-1904-08398} & $80.2$ & $67.6$\\
    CNN \cite{wang2019clinical}  & $77.0$ & $-$ \\
    CNN + Rules \cite{Yao2018ClinicalTC}  & $-$  & $96.2$\\
    
    \bottomrule
  \end{tabular}
\end{table}

We showcase the competitiveness of our architectures with respect to previous methods in Table \ref{tab:results}. In both shared tasks, top submissions consisted mainly of hand created regular expression and rules for each label. The top performing system in I2B2 2006 \cite{clark2008identifying} utilized handcrafted regular expressions, rules and feature sets to train per-label binary support vector machines.
The top performing system in I2B2 2008 \cite{ware2009natural} consisted of purely hand engineered rules for each label. As a simple baseline, we report the performance of predicting the majority occurring label across all training instances at evaluation. DocBert does not outperform either baseline. This is because it utilizes only the first 510 document tokens hence cannot capture any label indicating signal further into the document. For the I2B2 2006 dataset, a recent system \cite{wang2019clinical} explored a CNN architecture with word2vec representations but did not outperform the majority label or shared task baseline. Similarly, \cite{Yao2018ClinicalTC} utilized a CNN with word2vec alongside handcrafted features to achieve state of the art performance on I2B2 2008.

Our approach achieves state-of the art performance on smoking.
The $f_{\text{Transformer}}$ architecture failed to outperform most baselines across both tasks. In both tasks, $f_{\text{mean}}$, the simplest architecture, performed best. 
To the best of our knowledge, $f_{\text{mean}}$ beats the state of the art on I2B2 2006 by $2.8\%$.

The multi-head attention based Transformer encoder architecture $f_{\text{Transformer}}$ fails to learn a useful note representation across both tasks. We hypothesize this lower than expected performance is associated with the increased model capacity introduced by the architecture relative to the low number of training instances. Additionally, the utilization of multi-head attention results in the loss of temporal information amongst document sub-chunks (ex. certain document sections always precede others) which may contribute to the observed performance reduction.

The mean pooling architecture outperforms more complex parameterizations on both tasks. This suggests that the contextualized representation produced by BERT carries sufficient trait related signal through noise inducing pooling operations to dismiss the need for more complex architectures over the encoder.



\vspace{-.3cm}
\section{Conclusion}
We explore and contribute several document classification architectures that combine representations from state of the art language models. We find that treating document encoding as a sequence modeling task over sequential, contextualized document chunks is an effective framework for document representation agnostic of the classification architecture. All of our proposed architectures perform competitively with previous task baselines. Notably, we beat state of the art on a well known smoking status phenotyping task and demonstrate that simple strategies such as mean pooling are sufficient for training BERT based long document classifiers. We make our Pytorch implementation publicly available \footnote{\url{https://github.com/AndriyMulyar/bert\_document\_classification}}.

\small

\bibliography{updated_main_arxiv}
\bibliographystyle{plainnat}

\end{document}